\documentclass[11pt]{article}

\usepackage[final]{acl}
\usepackage{times}
\usepackage{latexsym}

\usepackage{amsmath,amssymb}
\usepackage{graphicx}
\usepackage{booktabs}
\usepackage{multirow}

\usepackage[T1]{fontenc}
\usepackage[utf8]{inputenc}
\usepackage{microtype} 
\usepackage{tabularx, colortbl}
\usepackage{framed}     
\usepackage{ragged2e}   
\usepackage{enumitem}   

\title{RSMeM: Knowledge-Enhanced Memory Evolution for Remote Sensing Agents with Systematic Evaluation\\[0.45em]
{\normalsize\normalfont
\href{https://github.com/AI9Stars/RSMeM}{%
\raisebox{-0.14em}{\includegraphics[height=1.05em]{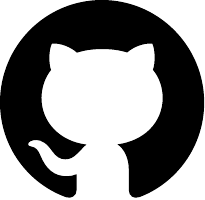}}\hspace{0.25em}\textbf{Code: GitHub}}
\quad\textbullet\quad
\href{https://modelscope.cn/studios/wbx929/RSMeM}{%
\raisebox{-0.14em}{\includegraphics[height=1.05em]{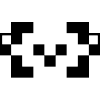}}\hspace{0.25em}\textbf{Demo: ModelScope Studio}}
}}

\author{
\textbf{Bingxian Wu\textsuperscript{1,3}\thanks{Equal contribution}},
\textbf{Yu Zhang\textsuperscript{2}\footnotemark[1]},
\textbf{Zonghao Guo\textsuperscript{2}\thanks{Corresponding authors}},
\textbf{Tang Liu\textsuperscript{1}},
\textbf{Chen Qian\textsuperscript{1,3}}, 
\textbf{Yuxiang Lu\textsuperscript{4}},\\
\textbf{Xingbo Du\textsuperscript{6}},
\textbf{Yanghao Li\textsuperscript{2}},
\textbf{Yidan Zhang\textsuperscript{3,5}}, 
\textbf{Chi Chen\textsuperscript{2}},
\textbf{Ling Yao\textsuperscript{3}},
\textbf{Maosong Sun\textsuperscript{2}}\\
\textsuperscript{1}Institute of Geographic Sciences and Natural Resources Research, CAS \quad
\textsuperscript{2}Tsinghua University \\
\textsuperscript{3}University of Chinese Academy of Sciences \quad
\textsuperscript{4}China University of Geosciences, Beijing \\
\textsuperscript{5}Aerospace Information Research Institute, CAS \quad
\textsuperscript{6}Shanghai Jiao Tong University \\
\texttt{wubingxian24@mails.ucas.edu.cn, guozonghao96@outlook.com}
}

\begin{document}
\maketitle
\begin{abstract}

Geoscience research requires complex analysis and domain expertise, with remote sensing (RS) observations as a key foundation.
However, existing RS agents built on general-purpose LLMs remain largely domain-agnostic, resulting in brittle and error-prone workflows. Moreover, these failures are seldom consolidated into a reusable experience for subsequent analyses. 
To address this issue, we introduce RSMeM, a knowledge-enhanced memory evolution mechanism that bootstraps RS agents with pre-distilled domain knowledge and iteratively integrates online experience for robust multi-step tool execution.
RSMeM is composed of two components: (i) Hierarchical Knowledge Grounding, which performs taxonomy-aware retrieval over a hierarchical domain corpus to guide planning and tool selection; and (ii) Failure-Aware Experience Refinement, which distills failure-annotated tool-use traces into reusable constraints for next-round tool execution. By iteratively employing these two processes, RS agents can evolve to absorb task-level domain knowledge and effectively translate it into instance-level execution experience. 
Extensive experiments on EarthBench demonstrate that RSMeM consistently improves tool-use performance and end-to-end answer across a diverse set of LLM backbones. Notably, {RSMeM} achieves a 6\% accuracy improvement on DeepSeek-V3.2 with less than 1\% additional experience tokens, demonstrating the strong knowledge density of our distilled experience.

\end{abstract}

\begin{figure}[t] 
  \centering
  \includegraphics[width=1.0\linewidth]{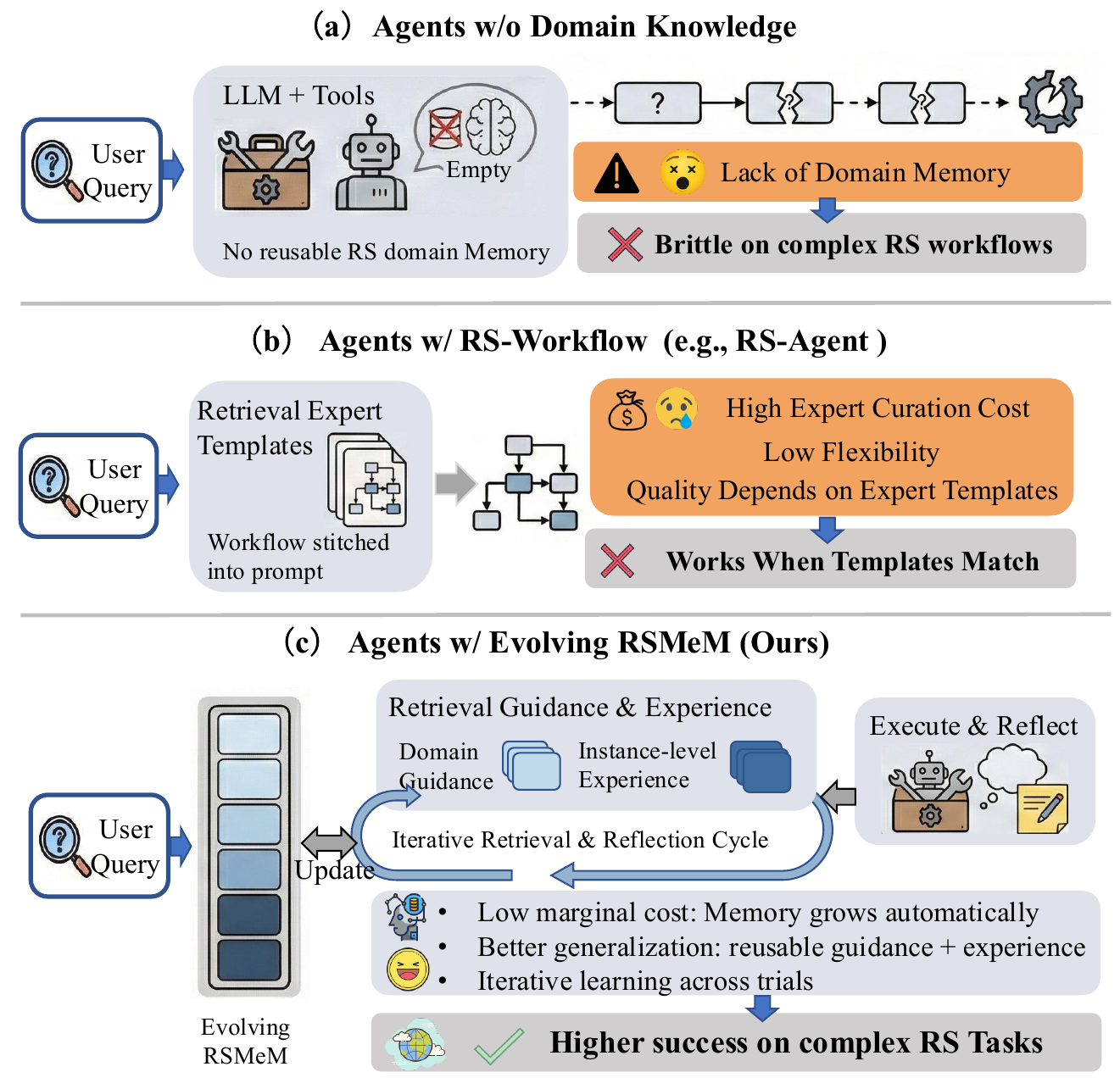}
  
  \caption{Comparison of RS agent architectures. (a) Agents w/o Domain Knowledge lack reusable memory, leading to brittle workflows. (b) Agents w/ RS-Workflow rely on static expert templates, which are inflexible and costly to curate. (c) Agent w/ Evolving RSMeM (Ours) integrates hierarchical grounding with failure-aware reflection, enabling iterative knowledge evolution and robust execution for complex RS tasks.}
  \label{fig:Overview}
\end{figure}

\section{Introduction}

Advances in large language models (LLMs) and tool-augmented agents have demonstrated promising capabilities in task planning, reasoning, and external tool use~\cite{su-etal-2024-language,tang2025agentkbleveragingcrossdomain}. 
In parallel, geoscience research has witnessed growing interest in deploying RS agents~\cite{xu2025rsagentautomatingremotesensing,shabbir2025thinkgeoevaluatingtoolaugmentedagents,feng2025earthagentunlockinglandscapeearth} for in-depth analysis of remote sensing imagery.

One line of RS agents~\cite{shabbir2025thinkgeoevaluatingtoolaugmentedagents,feng2025earthagentunlockinglandscapeearth} leverages general-purpose LLMs with ReAct-style tool-use to perform multi-step analysis. 
Yet, because these agents are largely domain-agnostic, they often struggle with domain-specific concepts (e.g., NDVI, BSI or land-surface temperature interactions), which makes their tool-use trajectories brittle and prone to cascading errors, such as premature termination or incorrect tool invocation, shown in Fig.~\ref{fig:Overview}(a). 
The other paradigm adopts semi-automated, expert-authored solution templates or workflow programs~\cite{xu2025rsagentautomatingremotesensing,chen2025canglingknowflowunifiedknowledgeandflowfusedagent} to guide LLMs planning and tool-calling, as shown in Fig.~\ref{fig:Overview}(b). 
While effective for narrowly scoped tasks, such workflow-centric designs are labor-intensive to construct and maintain. Moreover, these task-specific procedural details largely serve as external references, rather than enabling the agent to internalize corrections from execution failures and improve its behavior over time.

To enable agents to learn from mistakes on concrete tasks, prior work has explored reflection-driven improvements ~\cite{shinn2023reflexionlanguageagentsverbal,zhang2025agentlearningearlyexperience,fang2025mempexploringagentprocedural,zhang2025memevolvemetaevolutionagentmemory}. 
However, in geoscience settings, reflection alone is limited, as agents without domain knowledge often misdiagnose failures and fail to turn errors into procedural fixes.

To tackle this challenge, we present RSMeM, a memory-evolution framework that seeds RS agents with pre-distilled domain knowledge and continually incorporates online experience to enable robust multi-step tool execution, shown in Fig.~\ref{fig:Overview}(c).
Following prior task taxonomies in geoscience~\cite{KUCHARCZYK2021112577,MA2024113924,YUAN2020111716,MA2021103858}, we construct a set of geoscience knowledge corpus that covers 3 application domains, 12 subdomains and 64 task formulations. 
We further refine the corpus with manual, professional feedback to form a distilled knowledge base for bootstrapping RSMeM.
\begin{figure*}[t] 
  \centering
  \includegraphics[width=1.0\linewidth]{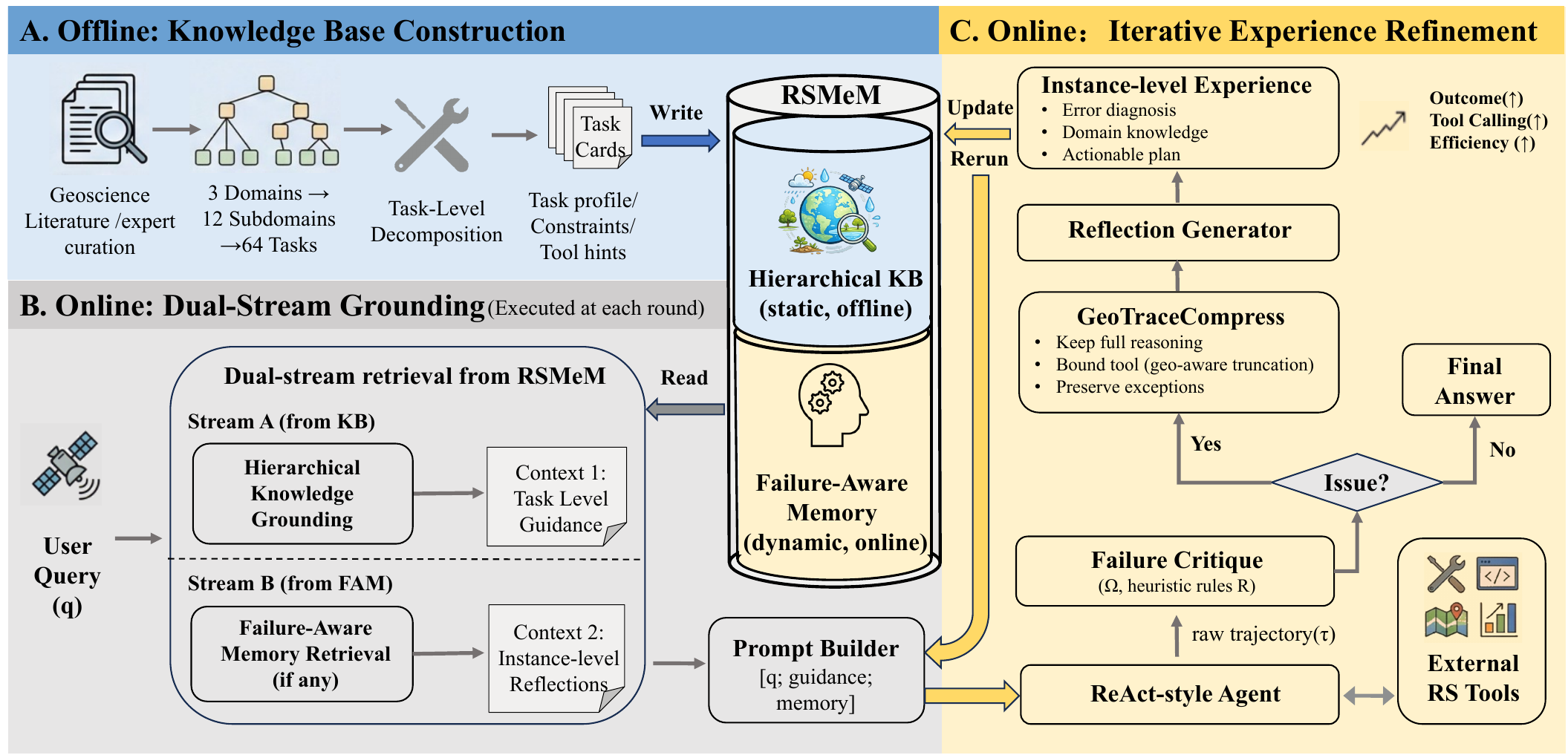}
  
  \caption{Overview of the RSMeM framework. The architecture consists of three main stages: (1) Offline Knowledge Base Construction: Systematically distilling geoscience literature into a hierarchical domain knowledge base and task-specific cards. (2) Online Task Solving via Dual-Stream Retrieval: The RS-Agent processes user queries and RS observations by retrieving task-level guidance through Hierarchical Knowledge Grounding (Stream A) and instance-level reflections from the Failure-Aware Memory (FAM) (Stream B). (3) Iterative Self-improvement: Based on the tool-use trace, a reflection generator synthesizes feedback to update the FAM, enabling the agent to evolve its experience and achieve higher accuracy and more reliable trajectories in subsequent tasks.}
  \label{fig:pipline}
\end{figure*}
RSMeM comprises two processes. 
\textbf{(i) Hierarchical Knowledge Grounding} conducts taxonomy-aware retrieval over a hierarchical corpus by encoding each knowledge item into retrievable units (keywords, task definitions, and semantic embeddings) and conditioning retrieval on the user query to inform LLM planning and tool selection. 
\textbf{(ii) Failure-Aware Experience Refinement} leverages a failure-critique environment to identify erroneous tool-use trajectories and distill them into reusable constraints that are stored as experience memory. 
By iterating between knowledge grounding and experience refinement, RSMeM injects both retrieved knowledge and failure-derived constraints into subsequent planning and tool invocation, enabling RS agents to progressively acquire instance-level execution competence from task-level domain knowledge.
Extensive experiments on EarthBench~\cite{feng2025earthagentunlockinglandscapeearth} demonstrate that RSMeM consistently improves tool-use performance and end-to-end answer quality across diverse LLM backbones, including DeepSeek V3.2~\cite{deepseekai2025deepseekv3technicalreport}, Kimi-K2~\cite{kimiteam2025kimik2openagentic}, Qwen3-32B~\cite{yang2025qwen3technicalreport} and Qwen3-8B~\cite{yang2025qwen3technicalreport}. 
Notably, {RSMeM} yields a 6.07\% absolute accuracy improvement on DeepSeek-V3.2 with less than 1\% additional experience tokens, whose efficiency arises from its ability to distill raw execution traces into high-density, reusable geoscience expertise. Rather than relying on resource-intensive context expansion, {RSMeM} prioritizes the synthesis of optimized workflow experiences, ensuring a superior performance-to-cost ratio for complex earth observation tasks.

We further observe that stronger backbones benefit more from RSMeM, as they exhibit greater capacity to summarize domain knowledge and diagnose execution errors, surfacing domain-level corrections during iterative refinement.

We summarize our contributions as follows:

\begin{itemize}
\item We propose a taxonomy-driven construction and retrieval scheme for structured geoscience knowledge, enabling RS agents to ground their reasoning in domain taxonomy thereby perform more reliable geoscience inference.
\item We propose a knowledge-enhanced memory evolution framework to iteratively couple knowledge grounding with failure-aware experience refinement, enabling RS agents to transform domain knowledge into reusable execution experience.
\item We propose RSMeM, a plug-and-play framework that can be seamlessly applied to diverse LLM backbones and consistently improves both tool-use and end-to-end performance on challenging remote sensing tasks.
\end{itemize}

\section{Related Works}
\textbf{Remote Sensing Agents.} Recently, agent-based Earth observation research has gained significant attention~\cite{luo2025geoevolveautomatinggeospatialmodel,kim2025sciencescalingagentsystems}. Tool-augmented agents excel in tasks such as code generation~\cite{huang2024agentcodermultiagentbasedcodegeneration}, multi-app collaboration~\cite{NEURIPS2024_0520537b}, and video understanding~\cite{fan2024videoagentmemoryaugmentedmultimodalagent}; however, their application in Earth observation remains in its infancy~\cite{kao2025llmagentsearthobservation}.
Early works, such as Change-Agent~\cite{Liu_2024}, focused on bi-temporal remote sensing change detection. In contrast, RS-ChatGPT~\cite{guo2024remotesensingchatgptsolving} and RS-Agent~\cite{xu2025rsagentautomatingremotesensing} integrated large language models with remote sensing tools for scene classification and object detection. Recent methods like ThinkGeo~\cite{shabbir2025thinkgeoevaluatingtoolaugmentedagents} introduced geospatial computing workflows, and UnivEarth~\cite{kao2025univearth} incorporated geo-environmental encoding for spectral analysis; however, high failure rates limit their practical use.
To overcome the overreliance on RGB data and underutilization of spectral information, Earth-Agent~\cite{feng2025earthagentunlockinglandscapeearth} integrates multispectral and hyperspectral data into a unified multimodal pipeline, supporting complex tasks like parameter inversion and time-series analysis, along with the Earth-Bench benchmark for evaluation. For structured geospatial workflows, the HTAM framework~\cite{li2025designingdomainspecificagentshierarchical} employs a task-aware hierarchical architecture to enable multi-step problem decomposition, instantiated as EarthAgent and evaluated through GeoPlan-Bench with metrics for tool selection and logical consistency. The CangLing-KnowFlow ~\cite{chen2025canglingknowflowunifiedknowledgeandflowfusedagent} introduces a Process Knowledge Base (PKB) and an evolutionary memory module to mitigate hallucinations.\\
\textbf{Memory-Augmented Experience Learning.} 
Memory enables language agents to retain and reuse information across interactions, substantially improving adaptability and task performance~\cite{fang2025mempexploringagentprocedural,xu2025amemagenticmemoryllm}. 
Prior work has explored various memory designs inspired by human cognition, aiming to support coherence, personalization, and continual learning in LLM-based agents~\cite{chhikara2025mem0buildingproductionreadyai}. Existing approaches can be broadly grouped into end-to-end memory systems, external memory mechanisms, and hierarchical memory architectures~\cite{hu-etal-2025-hiagent,tang2025agentkbleveragingcrossdomain}. Across these paradigms, experiences are typically stored as textual summaries or embedding representations and retrieved via semantic similarity, with memory update and forgetting strategies used to maintain relevance~\cite{shinn2023reflexionlanguageagentsverbal}. Memory is closely tied to experience learning, where agents enhance decision-making through repeated interactions with the environment~\cite{10.1145/3701716.3715247,fang2025mempexploringagentprocedural,zhang2026expseek}. While experience-driven learning has been explored through reinforcement and imitation learning~\cite{cai2025buildingselfevolvingagentsexperiencedriven,10610948}, general-purpose agents lack RS-specific frameworks that unify offline expertise with online instance-level experiences. We bridge this gap with {RSMeM}, which distills execution errors into actionable constraints to ground reasoning in a progressively refined, domain-specific memory.

\begin{table*}[t]
  \centering
  \footnotesize
  \caption{Tool-use Performance Comparison on Trajectory and Outcome.
  Best results per backbone are \textbf{bolded}; ``$\uparrow$'' indicates improvement over EarthAgent (R@3).}
  \label{tab:performance}

  \setlength{\tabcolsep}{4pt} 
  \renewcommand{\arraystretch}{1.1} 

  \begin{tabular}{p{1.9cm} p{2.1cm} p{1.0cm} c c c c c}
    \toprule
    \multirow{2}{*}{\textbf{Backbone}} &
    \multirow{2}{*}{\textbf{Method}} &
    \multirow{2}{*}{\textbf{Cfg.}} &
    \multicolumn{4}{c}{\textbf{Tool Calling Accuracy}} &
    \textbf{Outcome} \\
    \cmidrule(lr){4-7}
     & & & \textbf{AO} & \textbf{IO} & \textbf{EM} & \textbf{WS} & \textbf{Acc.} \\
    \midrule

    \rowcolor[gray]{0.95}
    \multicolumn{8}{l}{\textit{Proprietary Models}} \\
    GPT-5 & EarthAgent & R@1 & 69.11 & 58.25 & 45.57 & -- & 65.59 \\
    Gemini-2.5 & EarthAgent & R@1 & 57.96 & 45.44 & 31.86 & -- & 54.66 \\
    GPT-4o & EarthAgent & R@1 & 65.73 & 50.70 & 46.17 & -- & 45.34 \\
    \midrule

    \rowcolor[gray]{0.95}
    \multicolumn{8}{l}{\textit{Open-Source Models}} \\
    Qwen3-Max & EarthAgent & R@1 & 69.56 & 53.28 & 37.02 & -- & 50.20 \\
    LLaMA-4 & EarthAgent & R@1 & 16.51 & 2.45 & 1.70 & -- & 44.94 \\
    InternVL-3.5 & EarthAgent & R@1 & 8.83 & 3.87 & 2.02 & -- & 26.72 \\
    \midrule

    \rowcolor[gray]{0.95}
    \multicolumn{8}{l}{\textit{Main Evaluation}} \\

    \multirow{3}{*}{Qwen3-8B}
    & EarthAgent & R@1 & 43.91 & 16.68 & 3.79 & 0.0408 & 31.98 \\
    & EarthAgent & R@3 & 48.34 & 21.82 & 4.14 & 0.0455 & 39.27 \\
    & RSMeM & R@3 & \textbf{49.50} & \textbf{23.18} & \textbf{5.82} & \textbf{0.0600} & \textbf{42.11}$\uparrow$ \\

    \addlinespace[0.3em]
    \multirow{3}{*}{Qwen3-32B}
    & EarthAgent & R@1 & 38.66 & 14.75 & 5.42 & 0.0569 & 38.87 \\
    & EarthAgent & R@3 & 46.68 & 26.80 & 8.81 & 0.0930 & 44.94 \\
    & RSMeM & R@3 & \textbf{46.94} & \textbf{30.75} & \textbf{13.88} & \textbf{0.1469} & \textbf{48.58}$\uparrow$ \\

    \addlinespace[0.3em]
    \multirow{3}{*}{Kimi-K2}
    & EarthAgent & R@1 & 72.79 & \textbf{48.50} & 28.72 & 0.3975 & 36.44 \\
    & EarthAgent & R@3 & \textbf{73.39} & 48.47 & 29.12 & 0.3887 & 43.32 \\
    & RSMeM & R@3 & 70.88 & 47.41 & \textbf{30.81} & \textbf{0.3872} & \textbf{48.99}$\uparrow$ \\

    \addlinespace[0.3em]
    \multirow{3}{*}{DeepSeek-V3.2}
    & EarthAgent & R@1 & 77.27 & 55.02 & 26.56 & 0.4275 & 48.18 \\
    & EarthAgent & R@3 & 78.02 & 55.47 & \textbf{27.25} & \textbf{0.4342} & 51.82 \\
    & RSMeM & R@3 & \textbf{79.20} & \textbf{55.49} & 27.01 & 0.4231 & \textbf{57.89}$\uparrow$ \\
    \bottomrule
  \end{tabular}
\end{table*}

\section{Methodology}
\begin{figure}[t] 
  \centering
  \includegraphics[width=1.0\linewidth]{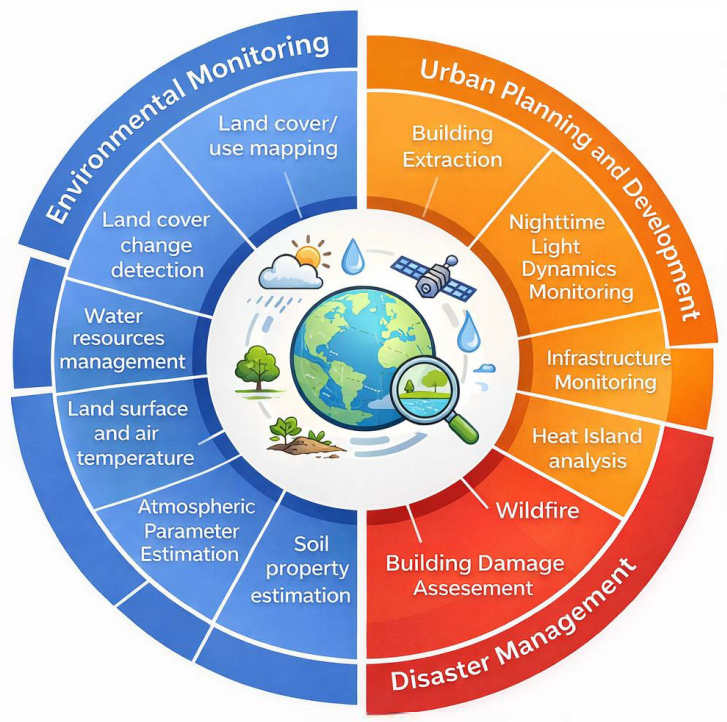}
  \caption{Hierarchical structure of the geoscience knowledge base.
The outer ring corresponds to high-level domains, and the inner ring captures fine-grained sub-domains.}
  \label{fig:domain_classifcation}
\end{figure}

As depicted in Fig.~\ref{fig:pipline}, {RSMeM} augments general LLMs with geoscientific expertise by coupling pre-distilled geoscience knowledge with iterative experience refinement. 
This iterative experience evolution follows a dual-stream mechanism that combines hierarchical knowledge grounding with experience refinement, which are operationalized via the prompt templates in Appendix~\ref{sec:appendix_prompts}.

\subsection{Hierarchical Knowledge Base Construction}
To provide a robust foundation for grounding, we construct a hierarchical geoscience knowledge base (KB) curated by domain experts. This expert-driven design ensures analytical validity and terminological consistency across complex RS tasks.
\paragraph{Hierarchical Taxonomy} The KB follows a three-level hierarchy consisting of 3 high-level \textit{Geoscience Domains}, which are further decomposed into 12 \textit{Sub-domains} and 64 \textit{Atomic Analytical Tasks}, as shown in Fig.~\ref{fig:domain_classifcation}. This structure allows the agent to navigate from broad application areas to specific operational steps.

\paragraph{Task Card Construction} Each atomic task is encapsulated in a \textbf{Task Card}, represented as a structured triplet $s_i = \{T_i, D_i, S_i\}$. Here, $T_i$ denotes the task title, $D_i$ provides a domain definition, and $S_i$ specifies the standardized suggestion or hints required for execution.

We provide comprehensive task card samples and details on the expert-driven construction process in Appendix \ref{sec:kb_construction} and \ref{sec:kb_samples}, respectively.

\subsection{Dual-Stream Online Grounding}

During inference, the agent assembles its input prompt from two complementary retrieval streams, appended after the system prompt and the user query:

\paragraph{Stream A: Hierarchical Knowledge Grounding (HKG).} Given a user query $q$, we retrieve the most relevant Task Card via a hybrid scoring function that combines neural similarity with taxonomy-aware lexical signals:
\begin{equation}
\begin{aligned}
\mathrm{Score}(q, s_i) =\; & \cos(\mathbf{v}_q, \mathbf{v}_i) + \alpha \cdot \mathbb{1}_{\mathrm{T}}(q, T_i) \\
& + \beta \cdot \textstyle\sum_{j=1}^{m} \mathbb{1}_{\mathrm{K}}(q, k_j) \\
& + \gamma \cdot \mathbb{1}_{\mathrm{C}}(q, S_i),
\end{aligned}
\end{equation}
where $\mathbf{v}_q$ and $\mathbf{v}_i$ are sentence-transformer embeddings of the query and the contextualized Task Card, respectively. The indicator terms $\mathbb{1}_{\mathrm{T}}$, $\mathbb{1}_{\mathrm{K}}$, and $\mathbb{1}_{\mathrm{C}}$ fire when query tokens lexically overlap with the task title $T_i$, curated domain keywords $\{k_j\}_{j=1}^m$, and the task-card content $S_i$, respectively. These lexical bonuses ($\alpha{=}0.2$, $\beta{=}0.08$, $\gamma{=}0.1$) compensate for cases where geoscience terminology (e.g., ``TVDI'', ``emissivity'') is underrepresented in the embedding space. We return the top-1 candidate exceeding a minimum threshold $\tau{=}0.2$.

\paragraph{Stream B: Failure-Aware Memory (FAM) Retrieval.}
For each instance $i$, the agent retrieves its most recent reflection from the FAM via a deterministic instance-indexed lookup:
\begin{equation}
c_i^{(t)} = \mathrm{Latest}\bigl(\mathcal{M}_i^{(t-1)}\bigr),
\end{equation}
where $\mathcal{M}_i^{(t-1)}$ is the reflection memory accumulated for instance $i$ up to round $t{-}1$, and $\mathrm{Latest}(\cdot)$ selects the most recent entry. This design is deliberate: by binding reflections to the specific execution trace that produced them, we ensure that corrective guidance is maximally precise and free from interference by unrelated failure patterns. Retrieved reflections provide negative constraints and historical fixes distilled from previous tool-use traces, preventing the repetition of execution errors (detailed in Appendix~\ref{sec:appendix_fam_samples}). 

\subsection{Iterative Experience Refinement}
The core of {RSMeM} is a self-improvement loop that progressively refines task-level knowledge into precise instance-level experience.

\paragraph{Failure Critique.}
A symbolic critic $\Omega$ (Fig.~\ref{fig:pipline}) applies a set of heuristic rules $\mathcal{R}$ to the agent's tool-use trajectory $a_i^{(t)}$ to determine whether the attempt should be considered failed:
\begin{equation}
\mathrm{Issue}(a_i^{(t)}) = \mathbb{I}\bigl(\exists\, \rho \in \mathcal{R} \;\text{s.t.}\; \rho(a_i^{(t)})\bigr),
\end{equation}
where each $\rho \in \mathcal{R}$ is a pattern-matching rule targeting a specific failure mode: (i)~missing or malformed answer tags, (ii)~empty responses, (iii)~invalid option letters, (iv)~refusal or incompleteness signals, and (v)~hedging language indicative of guessing. This lightweight symbolic critique operates without access to ground-truth labels.

\paragraph{GeoTraceCompress.}

Traces of RS agents often contain verbose geospatial I/O (e.g., lengthy file-path lists, large metadata dictionaries) that would overwhelm the reflection prompt. We apply a \textit{type-aware} compression strategy: list-valued outputs retain only the first three and last two items; dictionary outputs are reduced to two representative key-value pairs; error messages receive an elevated budget since they carry critical diagnostic signals. The agent's reasoning text is preserved in full. This compression reduces trajectory length by an order of magnitude while retaining the information most relevant for failure diagnosis.

\paragraph{Experience Distillation.}
For each failed instance, a \textbf{Reflection Generator} $R$ (instantiated as an LLM call with the same backbone) takes the original query $q_i$, the retrieved HKG context, and the compressed trajectory as input, and produces a structured reflection $r_i^{(t)}$---a 3--5 sentence diagnosis identifying the failure cause and proposing a corrective plan. The FAM is then updated:
\begin{equation}
\mathcal{M}_i^{(t)} = \mathcal{M}_i^{(t-1)} \oplus r_i^{(t)},
\end{equation}
where $\oplus$ denotes memory append. Once stored, this reflection becomes available to Stream~B for subsequent attempts, closing the execution--reflection--adaptation loop.

\begin{table}[t] 
\centering
\caption{Tool-use Performance Comparison (Efficiency). EarthAgent vs. memory-enhanced {RSMeM}. Best results per backbone are \textbf{bolded}; "↑" = improvement over EarthAgent (R@3).}
\label{tab:tool_performance_efficiency}
\footnotesize 
\setlength{\tabcolsep}{3pt} 
\renewcommand{\arraystretch}{1.1} 
\begin{tabular}{p{1.7cm} p{1.4cm} p{0.8cm} c c c}
\toprule
\multirow{2}{*}{\textbf{Backbone}} & \multirow{2}{*}{\textbf{Method}} & \multirow{2}{*}{\textbf{Cfg.} }&
\multicolumn{3}{c}{\textbf{Efficiency}} \\
\cmidrule(lr){4-6} 
& & & \textbf{TRI} & \textbf{Tok. (\%)} & \textbf{ED} \\
\midrule
\multirow{3}{1.7cm}{Qwen3-8B}
& EarthAgent & R@1 & 1.50 & 4.12 & -- \\
& EarthAgent & R@3 & \textbf{1.20} & 4.97 & -- \\
&  {RSMeM} &  R@3 &  1.26 &  \textbf{5.30} &  41.75 \\
\addlinespace[0.08em] 
\multirow{3}{1.7cm}{Qwen3-32B}
& EarthAgent & R@1 & \textbf{0.60} & 4.99 & -- \\
& EarthAgent & R@3 & 0.70 & 5.58 & -- \\
&  {RSMeM} &  R@3 &  0.65 &  \textbf{6.07} &  47.90 \\
\addlinespace[0.08em]
\multirow{3}{1.7cm}{Kimi-K2}
& EarthAgent & R@1 & 1.39 & 4.55 & -- \\
& EarthAgent & R@3 & 1.33 & 5.25 & -- \\
&  {RSMeM} &  R@3 &  \textbf{1.20} &  \textbf{6.02} &  48.81 \\
\addlinespace[0.08em]
\multirow{3}{1.7cm}{DeepSeek-V3.\allowbreak2}
& EarthAgent & R@1 & 2.03 & 5.99 & -- \\
& EarthAgent & R@3 & 1.92 & 6.23 & -- \\
&  {RSMeM} &  R@3 &  \textbf{1.86} &  \textbf{6.97} &  57.69\\
\bottomrule
\end{tabular}
\end{table}

\begin{table*}[t]
\centering
\caption{Ablation study of \textbf{HKG} (Hierarchical Knowledge Grounding) and 
\textbf{FAR} (Failure-Aware Experience Refinement) across representative backbones. 
RSMeM denotes the full integrated framework. \textbf{Cfg.} indicates the evaluation 
configuration (e.g., number of rounds). All metrics are aligned with the main results 
(Tab.~\ref{tab:performance}). Bold indicates the best performance within each backbone group.}
\label{tab:ablation_aligned_final}

\small
\setlength{\tabcolsep}{4pt}
\renewcommand{\arraystretch}{1.15}

\begin{tabular}{
p{2.2cm}  
p{1.0cm}  
c c
c c c c
c
c c c
}
\toprule
\multirow{2}{*}{\textbf{Variant}} & \multirow{2}{*}{\textbf{Cfg.}} &
\multicolumn{2}{c}{\textbf{Comp.}} &
\multicolumn{4}{c}{\textbf{Tool Calling Accuracy}} & 
\textbf{Outcome} &
\multicolumn{3}{c}{\textbf{Efficiency}} \\
\cmidrule(lr){3-4} \cmidrule(lr){5-8} \cmidrule(lr){9-9} \cmidrule(lr){10-12}
& & \textbf{HKG} & \textbf{FAR} &
\textbf{AO} & \textbf{IO} & \textbf{EM} & \textbf{WS} &
\textbf{Acc.} &
\textbf{TRI} & \textbf{Tok. (\%)} & \textbf{ED} \\
\midrule

\rowcolor[gray]{0.95}
\multicolumn{12}{l}{\textit{Backbone: Qwen3-8B}} \\
Baseline & R@1 & -- & -- & 43.91 & 16.68 & 3.79 & 0.0408 & 31.98 & 1.50 & 4.12 & -- \\
w/ HKG   & R@1 & \checkmark & -- & 44.87 & 17.33 & 4.15 & 0.0461 & 34.41 & 1.48 & 4.42 & -- \\
w/ FAR   & R@3 & -- & \checkmark & 48.36 & 23.35 & \textbf{6.55} & \textbf{0.0763} & 39.68 & 1.19 & 4.95 & 39.32 \\
\textbf{RSMeM} & R@3 & \checkmark & \checkmark & \textbf{49.50} & \textbf{23.18} & 5.82 & 0.0600 & \textbf{42.11} & \textbf{1.26} & \textbf{5.30} & \textbf{41.75} \\
\midrule

\rowcolor[gray]{0.95}
\multicolumn{12}{l}{\textit{Backbone: DeepSeek-V3.\,2}} \\
Baseline & R@1 & -- & -- & 77.27 & 55.02 & 26.56 & 0.4275 & 48.18 & 2.03 & 5.99 & -- \\
w/ HKG   & R@1 & \checkmark & -- & \textbf{79.50} & \textbf{56.51} & 25.85 & \textbf{0.4279} & 51.82 & 1.99 & 6.44 & -- \\
w/ FAR   & R@3 & -- & \checkmark & 76.94 & 54.73 & 26.75 & 0.4150 & 53.44 & 1.92 & 6.45 & 53.24 \\
\textbf{RSMeM} & R@3 & \checkmark & \checkmark & 79.20 & 55.49 & \textbf{27.01} & 0.4231 & \textbf{57.89} & \textbf{1.86} & \textbf{6.97} & \textbf{57.89} \\
\bottomrule
\end{tabular}
\end{table*}

\section{Experiment}

\subsection{Experimental Setup}
\textbf{\textit{Datasets:}}
To evaluate our method, we employ Earth-Bench~\cite{feng2025earthagentunlockinglandscapeearth}, a specialized benchmark designed for tool-augmented Earth Observation agents in the context of real-world geoscientific analysis. This benchmark comprises 248 expert-annotated instances spanning three distinct modalities: RGB, Spectrum, and Earth Products.\\
\textbf{\textit{Evaluated Models:}}
We evaluate RSMeM with four representative LLM backbones—DeepSeek-V3.2~\cite{deepseekai2025deepseekv3technicalreport}, Kimi-K2~\cite{kimiteam2025kimik2openagentic}, and Qwen3 (32B / 8B)~\cite{yang2025qwen3technicalreport}—and compare against EarthAgent under the same protocol. All results are aggregated over three independent rounds (R@3), where instances that remain unsolved are re-run in subsequent rounds, and the final accuracy is computed by backward overwriting with the latest successful attempt. 

\textbf{\textit{Reflexion Baseline:}}
We employ Reflexion~\cite{shinn2023reflexionlanguageagentsverbal} as a gradient-free baseline that iteratively refines performance by storing verbal self-reflections in episodic memory. To bridge the gap between its original Docstore design and our tool-augmented EO setting, we adapt its feedback loop to our environment's specific failure modes and structured output requirements. Detailed implementation and prompt templates are provided in Appendix~\ref{app:reflexion}.
\begin{figure*}[t] 
  \centering
  \includegraphics[width=1.0\linewidth]{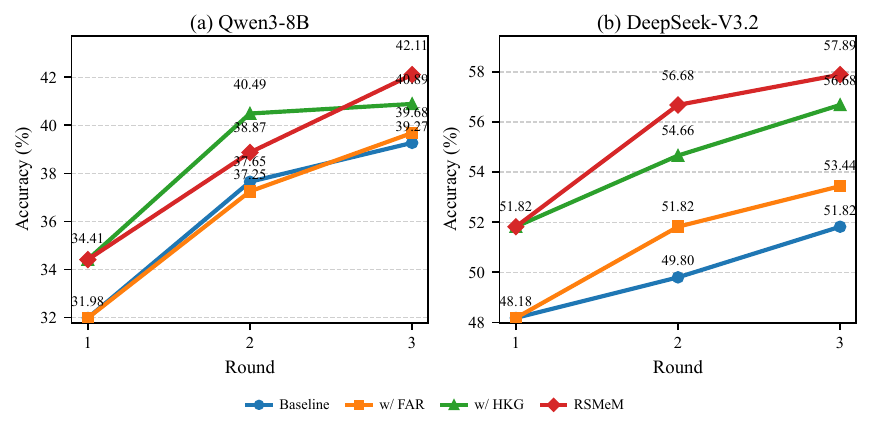}
  \caption{Ablation of RSMeM Components over Iterative Rounds (Accuracy across Models).}
  \label{fig:ablation_curvev2}
\end{figure*}

\textbf{\textit{Implementation Details:}}
Following EarthAgent's~\cite {feng2025earthagentunlockinglandscapeearth} configuration, we utilize 104 tools but substitute the MSCN model~\cite{10820553} with the SiamCRNN model~\cite{8937755}. Experiments are conducted on a remote sensing knowledge base across 3 application domains, 12 subdomains, and 64 tasks. We do not expose any ground-truth tool steps or answers to the agent during inference/refinement. Labels are reserved for evaluation only.
\subsection{Evaluation Metrics}
\label{sec:metrics}
We evaluate RSMeM across three systematic dimensions:
(i) Tool Calling Accuracy: Any-Order (AO), In-Order (IO), and Exact-Match (EM) for sequence matching, plus Wise-Score (WS) for prefix-sensitive trajectory and parameter precision;
(ii) Outcome: End-to-end Accuracy (Acc);
(iii) Efficiency: Trajectory Redundancy Index (TRI), Token Efficiency (Tok.), and Experience Density (ED) to quantify the cost-performance trade-off.
Formal definitions are provided in Appendix \ref{sec:appendix_eva}.

\section{Experimental Results}
\subsection{Main Results and Analysis}
We evaluate RSMeM in three dimensions (Tab.~\ref{tab:performance}, \ref{tab:tool_performance_efficiency}): tool-use correctness, outcome accuracy and efficiency.
\subsubsection{Tool-use Correctness across Granularities}
As shown in Tab.~\ref{tab:performance}, RSMeM consistently improves tool-use performance across AO, IO, EM, and WS, indicating gains at multiple levels of trajectory fidelity.
The improvement in AO suggests that RSMeM enhances global tool coverage by grounding planning in taxonomy-aware domain knowledge, while gains in IO reflect improved sequencing consistency. More importantly, the consistent increase in EM demonstrates that RSMeM reduces compounding errors across steps, enabling the agent to complete entire tool trajectories without deviation.
The rise in WS further confirms that these improvements are not limited to tool identity or order, but extend to parameter-level correctness under a prefix-sensitive evaluation. Since WS only rewards steps that follow a fully correct prefix, the observed gains imply that RSMeM stabilizes early-stage decisions, which are critical for downstream execution.
Tab.~\ref{tab:tool_performance_efficiency} evaluates the efficiency of RSMeM across TRI, token overhead, and ED. Compared to EarthAgent, RSMeM introduces a marginal increase in token consumption—for instance, rising from 6.23\% to 6.97\% on DeepSeek-V3.2, attributable to the structured reflection and memory update stages. However, this overhead remains strictly constrained across all backbones.

RSMeM consistently achieves high ED scores, indicating an effective balance between performance gains and computational cost. As backbone capacity increases, experience utility scales accordingly, suggesting that stronger models convert structured HKG grounding into more compact and actionable knowledge rather than redundant linguistic traces. Meanwhile, TRI values remain close to 1.0, showing that RSMeM avoids unnecessarily long or repetitive trajectories and instead concentrates computation on high-impact reflection, enabling accuracy improvements through refined planning rather than brute-force exploration.

We additionally compare RSMeM with the reflection-only baseline Reflexion.
As shown in Appendix~\ref{app:reflexion} (Fig.~\ref{fig:reflexion_compare}), RSMeM achieves consistent gains over Reflexion across all four backbones on both outcome accuracy and experience density, suggesting improvements beyond generic linguistic reflection. To further isolate the contribution of structured grounding from generic retrieval, we compare RSMeM against flat embedding retrieval and web-based augmentation (Appendix~\ref{sec:appendix_retrieval_baselines}); RSMeM outperforms both by a substantial margin (+16.33 pp over Simple RAG in accuracy), confirming that taxonomy-aware domain knowledge is fundamentally more effective than unstructured retrieval for geoscientific tasks.
A per-domain breakdown (Appendix~\ref{sec:appendix_domain_breakdown}) reveals that RSMeM benefits most in domains with standardized multi-step workflows (Disaster Management: +10.25 pp), while gains are smaller in domains where residual errors stem from upstream perception limits.

\subsubsection{Component ablation (HKG, FAR)}
Tab. \ref{tab:ablation_aligned_final} demonstrates that HKG and FAR provide complementary benefits to the tool-use process. HKG primarily optimizes global planning, increasing AO from 77.27\% to 79.50\% and IO from 55.02\% to 56.51\% on DeepSeek-V3.2. Conversely, FAR significantly enhances trajectory robustness, as evidenced by the WS improvement from 0.0408 to 0.0763 on Qwen3-8B. While individual components like HKG or FAR alone yield partial gains, their growth often stabilizes early or plateaus as shown in Fig. \ref{fig:ablation_curvev2}.

The full RSMeM framework achieves peak performance across all models, reaching outcome accuracies of 42.11\% and 57.89\% on Qwen3-8B and DeepSeek-V3.2, respectively. Despite a marginal increase in token usage, RSMeM consistently attains the highest ED, reaching 41.75 on Qwen3-8B and 57.69 on DeepSeek-V3.2—validating that the integration of both components ensures that performance gains are driven by high-density corrective knowledge rather than redundant computation.

\begin{table}[t]
\centering
\caption{Memory evolution on DeepSeek-V3.2. The baseline refers to the framework equipped with GRF}
\label{tab:memory_evolution_two_stage}
\footnotesize
\setlength{\tabcolsep}{3pt}
\renewcommand{\arraystretch}{1.15}

\begin{tabular}{lccccr}
\toprule
Setting & EM (\%) & Tok. (\%) & ED & Acc. (\%) & $\Delta$ \\
\midrule
GRF-only & 25.85 & 0.0644 & -- & 51.82 & -- \\
\midrule
\rowcolor[gray]{0.95} 
\multicolumn{6}{l}{\textit{Evolutionary Stages}} \\
Stage I: V1 & 26.19 & 0.0645 & 55.72 & 55.87 & {$\uparrow$ 4.05} \\
\textbf{Stage II: V2} & \textbf{27.01} & \textbf{0.0697} & \textbf{57.69} & \textbf{57.89} & \textbf{$\uparrow$ 6.07} \\
\bottomrule
\end{tabular}
\end{table}
\subsubsection{Memory Evolution and Iterative Effects}
Tab.~\ref{tab:memory_evolution_two_stage} further analyzes the evolutionary trajectory of memory refinement. Starting from a GRF-only baseline, EM increases from 25.85\% to 27.01\%, with outcome accuracy improving by +6.07 points after two evolution stages. Notably, the ED rises from 55.72 to 57.69. This upward trend suggests that memory evolution enhances "information purity," where subsequent stages distill coarse failure patterns into more concentrated, high-utility corrective knowledge rather than merely accumulating tokens. Fig.~\ref{fig:ablation_curvev2} visualizes this trend across iterative rounds. RSMeM exhibits steeper and more stable accuracy improvements than both the baseline and single-component variants on Qwen3-8B and DeepSeek-V3.2. While DeepSeek-V3.2 shows a rapid performance surge in the second round, followed by a marginal plateau, the integration of HKG and FAR enables improvements to persist where individual components (e.g., FAR-only) often stagnate. This indicates that structured grounding and failure-aware refinement reinforce each other, maintaining a high level of Experience Density even as accuracy gains stabilize.
Overall, these results demonstrate that RSMeM enhances tool use not by increasing trajectory length or trial-and-error, but by stabilizing early planning and converting execution failures into reusable experiences. This mechanism ensures a favorable effectiveness–efficiency trade-off, achieving consistent gains through superior experience density.

\begin{figure}[t]
  \centering
  \includegraphics[width=1.0\linewidth]{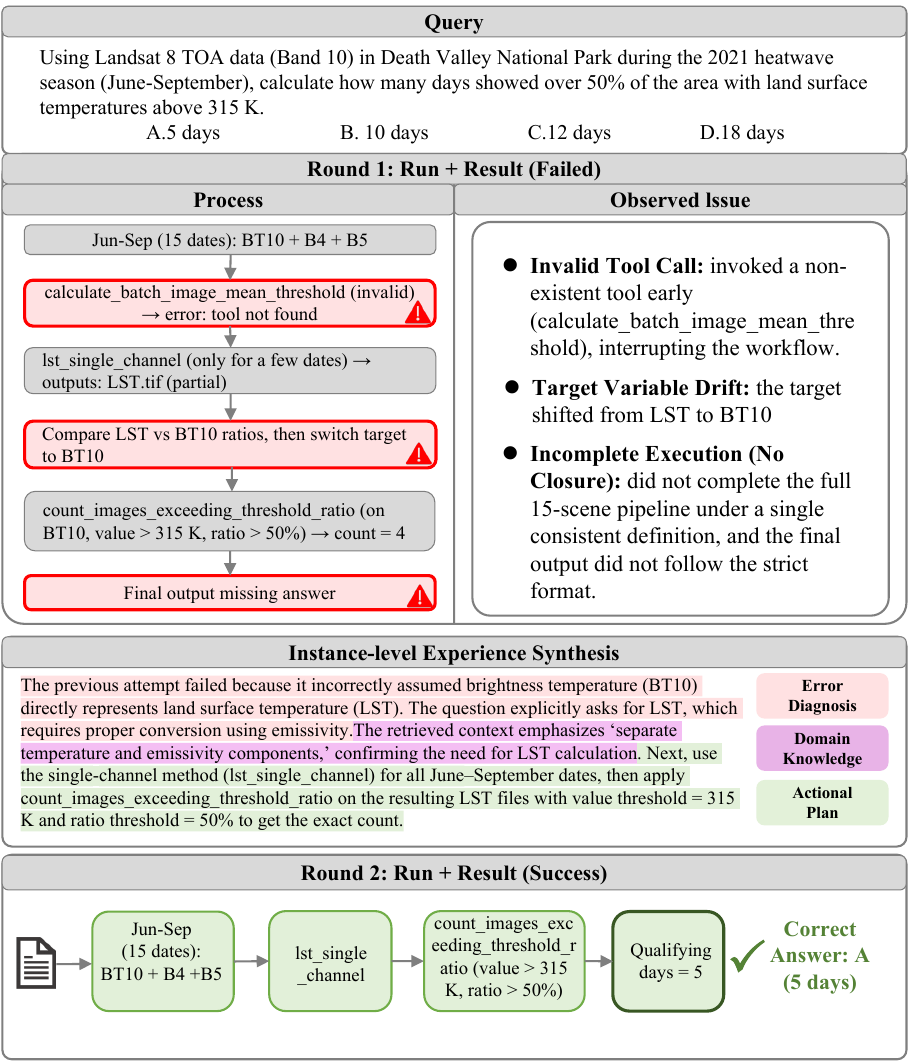}
  \caption{\textbf{Failure-aware memory evolution in Landsat thermal time-series analysis.} 
In Round1, the agent fails due to an invalid tool call and a semantic mismatch between BT10 (TOA brightness temperature from Landsat Band10) and LST (emissivity-corrected land surface temperature). RSMeM (DeepSeek-V3.2) grounds domain knowledge while diagnosing execution errors to distill instance-level experience, which subsequently guides a corrected end-to-end LST pipeline and yields the correct count of five days.}
  \label{fig:case_study}
\end{figure}

\subsection{Case Study}
Fig.~\ref{fig:case_study} illustrates a representative RS failure pattern in which errors are not isolated but instead cascade through the workflow. An early tool misuse, combined with a subtle semantic mismatch (e.g., treating BT10 as LST), jointly derails downstream steps, such that additional computation fails to recover correctness. This example highlights a key limitation of generic tool-using agents in geoscience tasks: they may execute syntactically plausible operations while violating the \emph{target variable contract} implicitly required by the task.

RSMeM improves robustness by converting such failures into persistent and reusable guardrails. Specifically, the experience refinement mechanism distills (i) a semantic constraint that enforces the correct variable definition (i.e., LST must be emissivity-aware), and (ii) a procedural constraint that maintains a consistent end-to-end pipeline with explicit completion requirements. The resulting behavioral change is not merely a repeated attempt, but a policy-level update that reduces the likelihood of recurring conceptual errors in subsequent thermal workflows. Overall, this case suggests that RSMeM’s performance gains primarily stem from stabilizing early commitments—namely, variable semantics and tool validity—thereby preventing error propagation and improving end-to-end reliability with minimal additional overhead.

\section{Conclusion}
We present RSMeM, a memory-evolution mechanism designed to bridge the gap between general-purpose LLMs and domain-specific geoscience requirements. By integrating Hierarchical Knowledge Grounding (HKG) and Failure-Aware Experience Refinement (FAR), RSMeM enables agents to bootstrap from pre-distilled expertise and iteratively optimize execution logic through online experience. Evaluations on EarthBench demonstrate that RSMeM consistently enhances tool-use accuracy and end-to-end performance across various LLM backbones. Notably, the framework achieves significant gains with minimal reflection overhead, establishing structured memory evolution as an efficient paradigm for building expert-level RS agents.
\section*{Limitations}
Despite its effectiveness, our work has several limitations. First, the performance of RSMeM in the initial stages is partially dependent on the quality and coverage of the hierarchical domain corpus; a sparse initial knowledge base may lead to a longer "cold-start" period for memory evolution. Second, while we evaluated the framework on the comprehensive EarthBench, its generalization to highly specialized or real-time streaming remote sensing tasks remains to be further explored. Lastly, the current experience refinement process primarily focuses on tool-use failures; incorporating successful but sub-optimal traces could potentially lead to even more nuanced strategic optimization, which we leave for future work.

\section*{Acknowledgments}
This work was supported in part by the National Natural Science Foundation of China under Grants 42501455 and 42471386, by the China Postdoctoral Science Foundation under Grant 2025M780333, and by the AI9Stars community.

\bibliography{custom}

\appendix

\section{Appendix}
\label{sec:appendix}
All artifacts are used under their respective open-source licenses for research purposes.
\subsection{Prompt Engineering and Agent Constraints} 
\label{sec:appendix_prompts}

The operational rigor and self-evolution capabilities of {RSMeM} are grounded in two core prompt templates. The \textit{System Prompt} ensures geoscientific professionalism and strict output formatting, while the \textit{Reflection Prompt} facilitates the distillation of failures into Failure-Aware Memory (FAM).

\subsubsection{System Prompt (Core Operational Logic)} 
The system prompt, as shown in Fig.~\ref{fig:system_prompt}, defines the agent's persona and enforces strict rules regarding tool usage and data path integrity, which are essential for multi-step Earth observation workflows.

\begin{figure}[ht] 
\small
\begin{framed}
\setlength{\parindent}{0pt} 
\RaggedRight 

\textbf{System Prompt Template} \\
\rule{\linewidth}{0.4pt} 

{\footnotesize \ttfamily \setlength{\emergencystretch}{3em}
\textbf{\#\#\# Core Role \& Instructions} \\
You are a professional geoscientist specializing in Earth observation data analysis. Follow these rules STRICTLY: \\
\textbf{1. Tool Usage:} Use designated tools to solve the problem step by step; retry ONLY ONCE if a tool returns an error. \\
\textbf{2. Path Rule:} When a tool returns ``Result saved at /path/to/file'', you must use the FULL path ``/path/to/file'' in all subsequent tool calls. \\
\textbf{3. Answer Format:} Output ONLY the correct choice in this format -- <Answer>Letter<Answer>. \\
\quad -- Do NOT add explanations or extra text outside the tags. \\
\quad -- Do NOT change the answer format under any circumstances.
}
\end{framed}
\caption{System Prompt Template for the geoscientific agent.} 
\label{fig:system_prompt}
\end{figure}

\subsubsection{Trajectory-Grounded Reflection Prompt (Stream B: Memory Generation)}
We adopt a \textit{Reflexion-style} self-critique prompt to distill failed tool-use trajectories into reusable, as shown in Fig.~\ref{fig:traj_reflection_prompt}, instance-specific constraints. Concretely, the prompt conditions the agent on (i) the original multiple-choice question (including data paths), (ii) optionally retrieved domain context, and (iii) a compressed chronological trajectory containing \texttt{tool\_use}, \texttt{tool\_result}, and key assistant decisions. The agent then produces a schema-constrained JSON reflection---diagnosing the failure and proposing an actionable, question-specific recovery plan---which is written into the Failure-Aware Memory (FAM, Stream B) for subsequent retrieval.

\subsubsection{Significance of the Dual-Prompt Strategy} 
As observed in our case studies Fig.~\ref{fig:case_study}, this dual-prompt strategy ensures that the agent is both a disciplined executor and a reflective learner. The \textit{System Prompt} maintains the precision required for LST retrieval, while the \textit{Reflection Prompt} enables the agent to autonomously correct methodological mismatches between retrieved knowledge and the specific task at hand.
\subsubsection{Reflection Prompt (Original Work)} 
This prompt, as shown in Fig.~\ref{fig:original_reflection_prompt}, represents the standard reflection mechanism used in the original ReAct-based frameworks, which relies on unstructured natural language for error diagnosis.

\begin{figure}[ht]
\small
\begin{framed}
\setlength{\parindent}{0pt}
\RaggedRight

\textbf{Trajectory-Grounded Reflection Prompt Template} \\
\rule{\linewidth}{0.4pt}

{\footnotesize 
You are an advanced reasoning agent that improves from self-reflection (Reflexion-style).\\
You will be given:
\begin{itemize}
\item The original question (includes data path and multiple-choice options),
\item Retrieved knowledge context (may be empty),
\item A compressed trajectory from the previous run ({\ttfamily tool\_use}, {\ttfamily tool\_result}, and key assistant text).
\end{itemize}
 
\textbf{Your task:} \\
In 3--5 sentences, diagnose the most likely reason for failure and devise a new, concise, high-level plan that avoids repeating the same failure. Focus on actionable, question-specific fixes.

\textbf{Required JSON schema:} \\
{\ttfamily \setlength{\emergencystretch}{3em}
\{ \\
\quad "question\_id": "\{question\_id\}", \\
\quad "reflective\_text": "..." \\
\}
}
}
\end{framed}
\caption{Trajectory-grounded reflection prompt template.}
\label{fig:traj_reflection_prompt}
\end{figure}

\begin{figure}[ht]
\small
\begin{framed}
\setlength{\parindent}{0pt}
\RaggedRight

\textbf{Reflection Prompt (Original Work)} \\
\rule{\linewidth}{0.4pt}

{\footnotesize
You are an advanced reasoning agent that can improve based on self-reflection. You will be given a previous unsuccessful reasoning trial and asked to analyze the failure.

In a few sentences, diagnose a plausible reason for failure and propose a concise, high-level mitigation plan.

\textbf{Previous trial:}\\
Question: {\ttfamily \{question\}}

\textbf{Reflection:}
}
\end{framed}
\caption{Baseline reflection prompt based on original Reflexion work.}
\label{fig:original_reflection_prompt}
\end{figure}

\subsubsection{Reflection Prompt (Modified)}
To ensure a fair and rigorous comparison, we modified the baseline prompt to align with our task-specific requirements, as shown in Fig.~\ref{fig:reflection_modified}. The Reflection Prompt (Modified) explicitly instructs the agent to diagnose failures related to tool usage, path integrity, and answer formatting, rather than relying on generalized error descriptions.

\begin{figure}[ht]
\small
\begin{framed}
\setlength{\parindent}{0pt}
\RaggedRight
\textbf{Reflection Prompt (Modified)} \\
\rule{\linewidth}{0.4pt}

{\footnotesize
\setlength{\emergencystretch}{5em}
You are an advanced reasoning agent that can improve based on self-reflection. You will be given a previous reasoning trial in which you had access to a tool environment and a question to answer. You were unsuccessful either because you produced an invalid final answer format, used refusal/guessing language, had data/path access issues, or used up your set number of reasoning steps.

In a few sentences, diagnose a possible reason for failure and devise a new, concise, high-level plan that aims to mitigate the same failure. Use complete sentences.

\textbf{Required JSON schema:} \\
\{ \\
\quad "question\_id": "\{question\_id\}", \\
\quad "reflective\_text": "..." \\
\}

\textbf{Previous trial:} \\
\{question\_text\}\{trajectory\_json\}

\textbf{Now output the JSON only:}
}
\end{framed}
\caption{The modified reflection prompt designed for structured comparison.}
\label{fig:reflection_modified}
\end{figure}

\subsection{Construction of the Hierarchical Knowledge Base}
\label{sec:kb_construction}

The hierarchical knowledge base (KB) in {RSMeM} is curated through a rigorous, expert-driven process designed to ensure analytical validity and geoscientific professionalism. Unlike purely automated or data-driven approaches, our KB prioritizes the formal logic of Earth observation research.

\paragraph{Expert Selection and Sources}
The curation was performed by a team of three domain experts with advanced degrees in Remote Sensing and Geospatial Information Science. The taxonomy and task definitions are grounded in authoritative sources, including the \textit{Remote Sensing Handbook} and recent state-of-the-art surveys in Earth observation.

\paragraph{Three-Tier Curation Process}
The experts followed a systematic top-down decomposition to ensure comprehensive coverage of the geoscience landscape:
\begin{itemize}
    \item \textbf{Domain Identification}: Three high-level domains (Urban Planning Development, Disaster Management, and Environmental Monitoring) were established to represent primary application areas.
    \item \textbf{Sub-Domain Mapping}: These were further decomposed into 12 sub-domains, such as Land Cover Mapping and Water Resources Management, ensuring semantic coherence.
    \item \textbf{Atomic Task Distillation}: A total of 64 atomic tasks were defined. Each task was mapped to a structured Task Card containing domain definitions ($D_i$) and expert-validated solutions ($S_i$), as exemplified in Tab.~\ref{tab:appendix_kb_detail}.
\end{itemize}
\paragraph{Lexicon and Validation}
To support the hybrid scoring in HKG, experts curated a deterministic keyword lexicon for each sub-domain, capturing both spectral properties and functional land-use categories. The final KB underwent a peer-review process within the expert group to eliminate semantic ambiguity and ensure that the retrieved tool-chains align with standard geoscientific practices. This expert-distilled expertise serves as the static "genetic" foundation for the subsequent online memory evolution.
\begin{table}[ht!]
\centering
\footnotesize 
\renewcommand{\arraystretch}{1.2}
\begin{tabularx}{\linewidth}{l X}
\toprule
\textbf{Task ID} & \textbf{Distilled Instance-level Memory (Reflective Text)} \\ \midrule
\rowcolor{gray!10} Q1 (TVDI) & \textbf{Failure:} Inefficient manual sampling and missing linear trend computation. \newline \textbf{Corrected:} Compute TVDI for all dates, aggregate to annual means (2019-2022), and compute regression slope. \\ \midrule
Q7 (LST) & \textbf{Failure:} Incorrectly used raw brightness temperature (BT10) instead of LST. \newline \textbf{Corrected:} Strictly apply single-channel LST algorithm (using NDVI emissivity) before threshold ratio analysis. \\ \midrule
\rowcolor{gray!10} Q9 (Urban) & \textbf{Failure:} Failed to define 'urban area' and missed emissivity correction. \newline \textbf{Corrected:} Calculate LST $\rightarrow$ Apply urban mask (NDVI < 0.2) $\rightarrow$ Count pixels $>$ 300K. \\ \midrule
Q12 (Spatial) & \textbf{Failure:} Processed isolated time points instead of daily composites. \newline \textbf{Corrected:} Compute daily LST maximums across all observations to capture peak heat stress ($>$30\%). \\ \bottomrule
\end{tabularx}
\caption{Examples of distilled Instance-level Memory (FAM).}
\label{tab:fam_examples}
\end{table}
\begin{table}[t]
\centering
\scriptsize 
\setlength{\tabcolsep}{3pt} 

\begin{tabularx}{\linewidth}{l X}
\toprule
\rowcolor[HTML]{DEDEDE}
\textbf{Level} & \textbf{Content / Metadata} \\ \midrule
\textbf{Domain} & \textbf{Environmental Monitoring} \\
\textbf{Sub-Domain} & \textbf{Land Cover Mapping} \\
\textbf{Keywords} & Residential, Commercial, Industrial, Recreational, Infrastructure, Land cover heterogeneity, Emissivity variation \dots \\ \midrule

\rowcolor[HTML]{F2F2F2}
\multicolumn{2}{l}{\textbf{Decomposed Atomic Tasks (Task Cards)}} \\ \midrule
\textbf{Task 1 ($T_1$)} & \textbf{Pixel-Wise Semantic Segmentation} \\
Sol. ($S_1$) & Target high-res thematic mapping by assigning semantic labels to pixels; leverages spectral signatures. \\ \midrule

\textbf{Task 2 ($T_2$)} & \textbf{Thermal-Based Heterogeneity} \\
Sol. ($S_2$) & Analyze thermal infrared data to quantify spatial variations; separate temperature and emissivity. \\ \midrule

\textbf{Task 3 ($T_3$)} & \textbf{Visual Scene Classification} \\
Sol. ($S_3$) & Automated recognition to identify and count specific scene types based on taxonomy. \\ \midrule

\textbf{Task 4 ($T_4$)} & \textbf{Sub-Pixel Spectral Unmixing} \\
Sol. ($S_4$) & Resolve material abundances by decomposing composite spectral signatures into endmembers. \\ \midrule

\rowcolor[HTML]{F2F2F2}
\multicolumn{2}{l}{\textbf{Domain Definitions}} \\ \midrule
\textit{$D_1$} & \textbf{Land cover classification}: Process of assigning spatial units to discrete categories. \\
\textit{$D_2$} & \textbf{Surface Emissivity}: Effectiveness of Earth's surface in emitting thermal radiation. \\ \bottomrule
\end{tabularx}
\caption{Detailed hierarchical structure of the RSMeM for the Land Cover Mapping sub-domain. The expert-curated domain knowledge is decomposed into hierarchical levels, with four representative atomic tasks shown out of the 64 available in our knowledge base.}
\label{tab:appendix_kb_detail}
\end{table} 
\subsection{Hierarchical Knowledge Base Detail}
\label{sec:kb_samples}
We provide a detailed decomposition of the \textbf{Environmental Monitoring} domain within our RSMeM, specifically focusing on the \textbf{Land Cover Mapping} sub-domain. This sample illustrates how expert-curated summaries are operationalized into discrete task cards $\{T_i, D_i, S_i\}$ to guide the agent.
\subsection{Instance-level Memory (FAM) Samples}
\label{sec:appendix_fam_samples}

Instance-level memory, denoted as \textit{Stream B} in our framework, consists of successful reasoning patterns and error-correction logs distilled from historical interactions. When an agent fails a task, a self-reflection module generates a \texttt{reflective\_text} that summarizes the failure cause and the corrected strategy. 

Tab.~\ref{tab:fam_examples} provides representative samples of how these reflections are stored and subsequently retrieved to guide future similar tasks.

\subsection{Evaluation Metrics Details}
\label{sec:appendix_eva}
The proficiency of Large Language Models (LLMs) in utilizing external tools is assessed through a comprehensive set of metrics. We separately report tool calling accuracy and task-level outcome with efficiency, as improvements in execution correctness do not necessarily translate to final answer accuracy or computational cost.

\paragraph*{Tool Calling Accuracy}
\label{par:tool_accuracy}
These metrics focus on the model's ability to correctly select and sequence the necessary tools based on the ground truth sequence of expected tools, $\mathrm{GT}_{\mathrm{Tools}} = [t_1, t_2, \dots, t_n]$. Each metric is calculated as the proportion of questions where the model's tool trajectory satisfies the specific criteria. Let $M_{\mathrm{Tools}}$ denote the set or sequence of tools called by the model, and $N$ be the total number of questions.

\textbf{\textit{Any-Order (AO):}} Measures the success rate where the \textbf{set} of tools called by the model ($M_{\mathrm{Tools}}$) exactly matches the set of expected tools, irrespective of execution order:
\begin{equation}
\mathrm{AO} = \frac{1}{N}\sum_{i=1}^{N} \mathbb{I} \{ \mathcal{S}(M_{\mathrm{Tools}}^i) = \mathcal{S}(\mathrm{GT}_{\mathrm{Tools}}^i) \}.
\label{eq:any-order}
\end{equation}

\textbf{\textit{In-Order (IO):}} Measures the success rate where the \textbf{sequence} of tools called by the model precisely matches the chronological sequence of the expected tools:
\begin{equation}
\mathrm{IO} = \frac{1}{N}\sum_{i=1}^{N} \mathbb{I} \{ \mathcal{Q}(M_{\mathrm{Tools}}^i) = \mathcal{Q}(\mathrm{GT}_{\mathrm{Tools}}^i) \}.
\label{eq:in-order}
\end{equation}

\textbf{\textit{Exact-Match (EM):}} Measures the success rate where the entire model-generated tool-calling trajectory is an identical sequence to the reference steps, including the total number of calls, the specific tools invoked, and their order:
\begin{equation}
\mathrm{EM} = \frac{1}{N}\sum_{i=1}^{N} \mathbb{I} \{ M^i = \mathrm{GT}^i \}.
\label{eq:exact-match}
\end{equation}
\begin{figure}[ht]
\small
\begin{framed}
\noindent \textbf{Agent Reasoning Trace (Stream A: HKG Guidance)} \\
\rule{\linewidth}{0.4pt} 

\noindent \textbf{User Query:} ...calculate how much cooler was the forested area (NDVI > 0.7) compared to the surrounding non-vegetated area (NDVI < 0.2)...

\noindent \fbox{
  \parbox{0.95\linewidth}{
    \footnotesize \textbf{--- HKG Context: Task-Level Guidance (Stream A) ---} \\
    \textbf{Domain:} Env. Monitoring \quad \textbf{Sub-domain:} Land Cover \\ 
    \textbf{Task:} Thermal-Based Heterogeneity Detection\\
    \textbf{Suggested:} Analyze thermal infrared data to quantify spatial variations; use single-channel LST retrieval with NDVI-based emissivity.
  }
}

\begin{itemize}[leftmargin=1.2em, noitemsep, topsep=4pt, label=\textbullet]
    \item \textbf{Step 1 (Tool: {\ttfamily get\_filelist})} \\
    \textit{Obs:} [{\ttfamily BT10.tif}, {\ttfamily b4.tif}, {\ttfamily b5.tif}]
    
    \item \textbf{Step 2 (HKG-Guided):} Apply single-channel method. \\
    \textbf{Tool:} {\ttfamily lst\_single\_channel(bt\_path=..., ...)} \\
    \textit{Obs:} Saved at {\ttfamily /path/to/\allowbreak lst\_result.tif} 
    
    \item \textbf{Step 3 (Thinking):} Mean LST for $NDVI > 0.7$ and $NDVI < 0.2$. \\
    \textbf{Tool:} {\ttfamily calc\_mean\_lst(..., mode='above')} $\rightarrow$ \textit{290.94K} \\
    \textbf{Tool:} {\ttfamily calc\_mean\_lst(..., mode='below')} $\rightarrow$ \textit{281.97K}
\end{itemize}

\hrule
 
\noindent \textbf{Final Answer:} \\
\textit{Thinking:} Forested (290.94K) vs. Non-veg (281.97K). Diff $\approx$ 8.97K, closest to \textbf{Option D}. \\
\textbf{Output:} {\ttfamily <Answer>D<Answer>}
\end{framed}
\caption{Full execution trace of the agent leveraging HKG (Stream A).}
\label{fig:trace_sample}
\end{figure}

\textbf{\textit{Wise-Score (WS):}}
Wise-Score jointly evaluates trajectory correctness and parameter-level accuracy in a prefix-sensitive and importance-aware manner.
It measures whether a model follows a correct sequence of tool invocations, where the contribution of each step is conditioned on all preceding steps being correct.
Moreover, each tool call is weighted to reflect its relative importance in the overall task.

Formally, the score is defined as:
\begin{equation}
\mathrm{WS} = \frac{1}{N} \sum_{i=1}^{M}
\left(
\mathbb{I}(\mathrm{Tool}_{\mathrm{act}}^{i} = \mathrm{Tool}_{\mathrm{gt}}^{i})
\cdot \gamma_i
\right),
\label{eq:wise_score}
\end{equation}

where $N$ denotes the total number of tool calls in the ground-truth trajectory, and $M$ is the length of the longest continuous matching prefix such that the predicted tool matches the ground truth at every step $j \le M$.
The term $\gamma_i$ represents the attribute-level correctness score of the $i$-th tool call, which serves as an importance-aware weight capturing the correctness and relevance of the associated parameters for that tool.
\paragraph*{End-to-End Accuracy}
\label{par:end_to-end}

This metric confirms successful synthesis of tool results to produce the correct overall output.
\begin{figure}[ht]
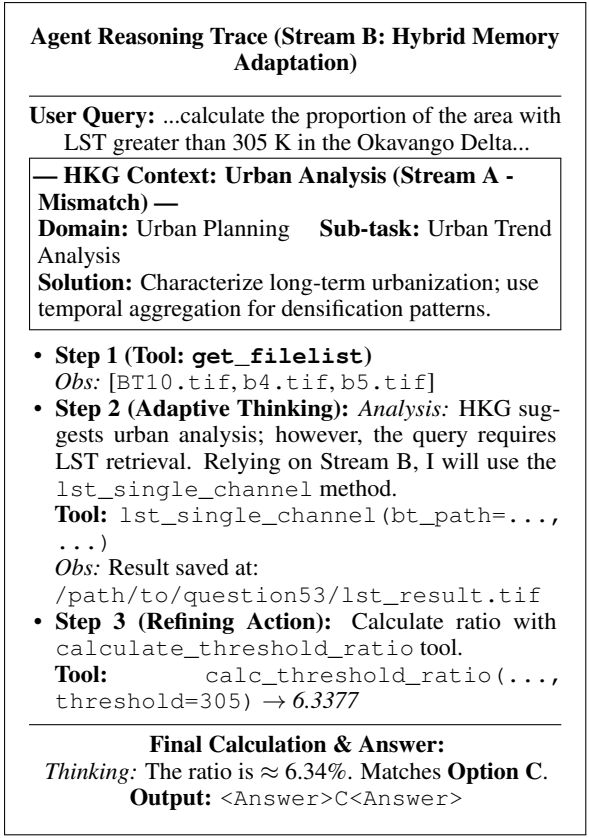

\small
\centering
\begin{framed}
\setlength{\parindent}{0pt} 
\textbf{Agent Reasoning Trace (Stream B: Hybrid Memory Adaptation)} \\
\rule{\linewidth}{0.4pt} 

\textbf{User Query:} ...calculate the proportion of the area with LST greater than 305 K in the Okavango Delta...

\noindent\fbox{%
  \begin{minipage}{\dimexpr\linewidth-2\fboxsep-2\fboxrule}
    \footnotesize \RaggedRight
    \textbf{--- HKG Context: Urban Analysis (Stream A - Mismatch) ---} \\
    \textbf{Domain:} Urban Planning \quad \textbf{Sub-task:} Urban Trend Analysis \\
    \textbf{Solution:} Characterize long-term urbanization; use temporal aggregation for densification patterns.
  \end{minipage}%
}

\begin{itemize}[leftmargin=1.1em, rightmargin=0.2em, noitemsep, topsep=4pt, label=\textbullet]
    \item \textbf{Step 1 (Tool: {\ttfamily get\_filelist})} \\
    \textit{Obs:} [{\ttfamily BT10.tif}, {\ttfamily b4.tif}, {\ttfamily b5.tif}]
    
    \item \textbf{Step 2 (Adaptive Thinking):} \textit{Analysis:} HKG suggests urban analysis; however, the query requires LST retrieval. Relying on Stream B, I will use the {\ttfamily lst\_single\_channel} method. \\
    \textbf{Tool:} {\ttfamily lst\_single\_channel(bt\_path=..., ...)} \\
    \textit{Obs:} Result saved at: \\
    {\footnotesize \ttfamily /path/to/question53/\allowbreak lst\_result.tif} 
    
    \item \textbf{Step 3 (Refining Action):} Calculate ratio with {\ttfamily calculate\_threshold\_ratio} tool. \\
    \textbf{Tool:} {\ttfamily calc\_threshold\_ratio(..., threshold=305)} $\rightarrow$ \textit{6.3377}
\end{itemize}

\hrule
\vspace{4pt}

\textbf{Final Calculation \& Answer:} \\
\textit{Thinking:} The ratio is $\approx$ 6.34\%. Matches \textbf{Option C}. \\
\textbf{Output:} {\ttfamily <Answer>C<Answer>}
\end{framed}
\caption{Full execution trace illustrating adaptive reasoning (Stream B).}
\label{fig:trace_mismatch}
\end{figure}
\textbf{\textit{Accuracy (Acc):}} The ultimate end-to-end performance measure, computed as the ratio of questions where the model's \textbf{Final Answer} ($y$) exactly matches the \textbf{Ground Truth} final answer ($y^*$):
\begin{equation}
\mathrm{Acc} = \frac{1}{N}\sum_{i=1}^{N} \mathbb{I} \{ y^i = y^{*i} \}
\label{eq:accuracy}
\end{equation}
where $\mathbb{I}\{\cdot\}$ is the indicator function that returns $1$ if the condition is met, and $0$ otherwise.
\paragraph*{Efficiency Metrics}
\label{par:efficiency_metrics}
These metrics evaluate the trade-off between the performance gains achieved through iterative reflection and the associated computational overhead.\\
\textbf{\textit{Trajectory Redundancy Index(TRI):}} Evaluates computational overhead by comparing the total number of steps in the model's trajectory to the minimum number of reference steps required. A value closer to $1.0$ indicates higher efficiency:
\begin{equation}
\mathrm{TRI} = \frac{\sum_{i=1}^{N} \mathrm{Steps}(M^i)}{\sum_{i=1}^{N} \mathrm{Steps}(\mathrm{GT}^i)}.
\label{eq:efficiency}
\end{equation}

\textbf{\textit{Token Efficiency (Tok.):}} To assess the cost-effectiveness of the autonomous evolution, we normalize the final success rate against the logarithmic total token expenditure. This represents the "accuracy yield" per unit of computational resource:
\begin{equation}
\eta = \frac{\mathrm{Acc}^{(T)}}{\log_{10}(\mathcal{C})}
\label{eq:token_efficiency},
\end{equation}
where $\mathcal{C}$ is the total cumulative token count consumed across all attempts, including inference tokens for $a_i^{(t)}$ and generation tokens for reflections $\mathcal{M}_i^{(t)}$. 

\textbf{\textit{Experience Density (ED):}} To further calibrate the effectiveness of the generated reflections under fixed hyperparameter constraints, we introduce Experience Density. Unlike the raw ratio, ED serves as a penalized utility metric that measures the ``information purity'' of the reflection trace:
\begin{equation}
\mathrm{ED} = \text{Acc} \times \left(1 - \frac{\mathcal{C}_{refl}}{\mathcal{C}_{total}}\right)
\label{eq:ed}
\end{equation}
where $\mathcal{C}_{refl}$ and $\mathcal{C}_{total}$ are predefined hyperparameters. ED quantifies the model's ability to compress corrective knowledge into a limited token window; a higher ED signifies that the hierarchical grounding effectively filters redundant linguistic fillers, thereby increasing the density of actionable experience.
\subsection{Qualitative Analysis and Case Studies}
\label{sec:trace_sample}

\paragraph{Validation of HKG Guidance.} The trace in Fig.~\ref{fig:trace_sample} exemplifies a successful execution driven solely by \textbf{Stream A (HKG)}. Even without prior instance-level experience from FAM (Stream B), the agent correctly identifies the professional tool-chain (\texttt{lst\_single\_channel} $\rightarrow$ \texttt{calculate\_mean\_lst\_by\_ndvi}) by grounding the user query into our hierarchical knowledge base. Specifically, the "Key Principle" retrieved from the Task Card prevents the agent from attempting generic image-to-temperature mappings, ensuring that geoscientific physical constraints are maintained throughout the multi-step reasoning process.
\textbf{Scenario.} The agent is tasked with calculating the Land Surface Temperature (LST) difference between forested (NDVI $>$ 0.7) and non-vegetated (NDVI $<$ 0.2) areas in the Black Forest region using Landsat 8 imagery, the analysis as shown in Fig.~\ref{fig:trace_mismatch}.

\paragraph{Resilience under Mismatched Guidance} 
The trace in Fig.~\ref{fig:trace_mismatch} demonstrates the robustness of {RSMeM} when \textbf{Stream A (HKG)} retrieves a sub-optimal Task Card. Despite the mismatch, the agent leverages its \textbf{Instance-level Memory (Stream B)} to maintain analytical rigor.

\textbf{Scenario.} The agent is tasked with calculating the proportion of the area with Land Surface Temperature (LST) greater than 305 K in the Okavango Delta using Landsat 8 data (Band 10, 4, and 5), the analysis as shown in Fig.~\ref{fig:trace_mismatch}.

\subsection{Comparison with Reflexion}
\label{app:reflexion}
We further compare RSMeM with Reflexion, Fig.\ref{fig:reflexion_compare} summarizes the Round@3 results across four LLM backbones.
\subsubsection{Implementation of the Reflexion Baseline}

To ensure Reflexion operates effectively within the EO environment, we specialized its feedback mechanism as follows:

Redefined Failure Signals: We expanded the original termination criteria to include environment-specific errors such as invalid answer formats, tool/path access failures, and maximum step exhaustion.

Lightweight Taxonomy: To minimize redundancy and token overhead, we replaced verbose textual reflections with a JSON-only schema based on a predefined failure taxonomy.

Prompt Adaptation: The original templates were restructured to prioritize tool-calling logic while maintaining the core reflection-loop integrity, as shown in Appendix~\ref{sec:appendix_prompts}.
\subsubsection{Detailed Analysis}
\begin{figure}
    \centering
    \includegraphics[width=1.0\linewidth]{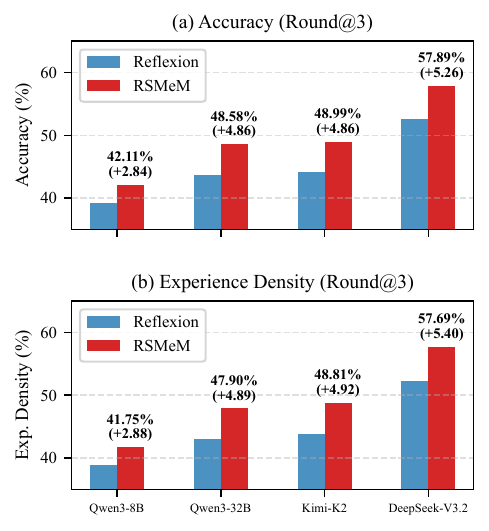}
    \caption{\textbf{Comparison with Reflexion under Round@3.} RSMeM consistently outperforms the reflection-only baseline Reflexion across four LLM backbones. (a) Outcome accuracy (\%) and (b) Experience Density (ED, \%) on Earth-Bench. Numbers above RSMeM bars report absolute scores, while values in parentheses denote the improvement over Reflexion.}
    \label{fig:reflexion_compare}
\end{figure}
Overall, RSMeM consistently outperforms Reflexion on both outcome accuracy and Experience Density (ED) across all evaluated backbones, as shown in Fig.~\ref{fig:reflexion_compare}. The improvements are stable rather than model-specific, indicating that RSMeM provides benefits beyond generic linguistic reflection by grounding decisions with structured domain knowledge and distilling failure-aware, reusable constraints.
Notably, ED improves alongside accuracy, suggesting that the gains are achieved through higher-utility experience utilization rather than increased trial-and-error or redundant computation.
\begin{table}[t]
\centering
\caption{Reflection-token overhead of RSMeM. We report the fraction of tokens consumed by the reflection stage in the entire inference budget, i.e., $\mathcal{C}_{\mathrm{refl}}/\mathcal{C}_{\mathrm{total}}$, where $\mathcal{C}_{\mathrm{refl}}$ counts tokens generated during the post-hoc reflection call and $\mathcal{C}_{\mathrm{total}}$ counts all tokens generated in an episode. Results correspond to the same RSMeM evaluation setup as Table~1 (R@3).}
\label{tab:reflection_token_ratio}
\footnotesize
\setlength{\tabcolsep}{4pt}         
\renewcommand{\arraystretch}{1.18}  
\begin{tabular}{@{}p{2.6cm}@{\hspace{1.2cm}}r@{}}
\toprule
\textbf{Backbone} & \textbf{Refl./Total (\%)} \\
\midrule
Qwen3-8B        & 0.85 \\
Qwen3-32B       & 1.40 \\
Kimi-K2         & 0.37 \\
DeepSeek-V3.2   & 0.35 \\
\bottomrule
\end{tabular}
\end{table}

\subsection{Reflection-Token Overhead of RSMeM}
As shown in Table~\ref{tab:reflection_token_ratio}, the reflection stage
introduces negligible token overhead across all tested backbones. The
fraction of reflection tokens remains below $1.5\%$ of the total
inference budget in every case, confirming that RSMeM's post-hoc
reflection mechanism is computationally lightweight. Notably, the smaller-scale backbones (Qwen3-8B and Qwen3-32B) incur slightly higher relative costs (0.85\% and 1.40\%, respectively), likely because their reflective outputs tend to be more verbose. In contrast, the stronger backbones, Kimi-K2 and DeepSeek-V3.2, consume only 0.37\% and 0.35\% of total tokens for reflection, suggesting that higher-capacity models generate more concise and focused reflective responses.

\begin{figure}[h]
  \centering
  \includegraphics[width=\columnwidth]{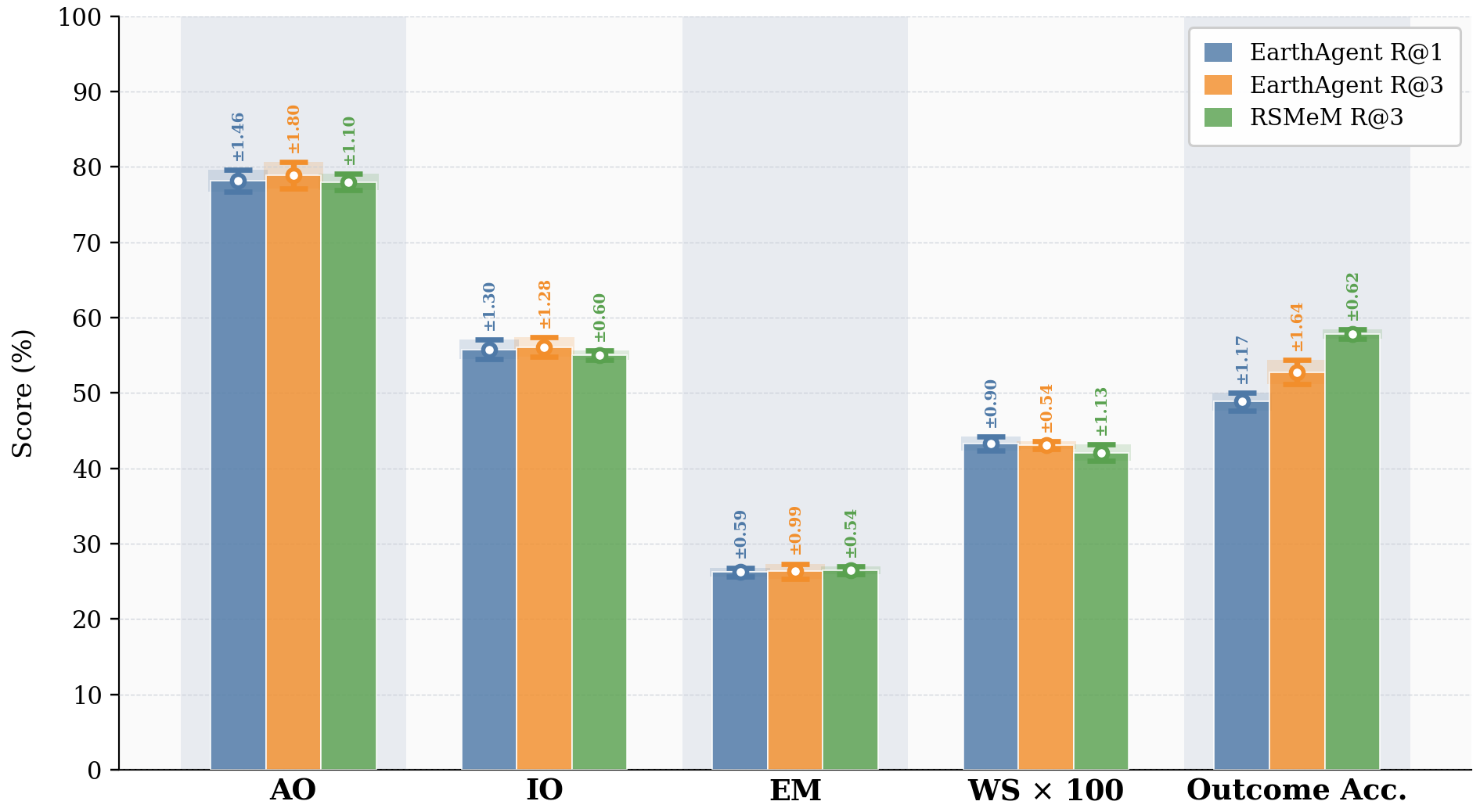}
  \caption{DeepSeek-V3.2 results (mean $\pm$ std, $n{=}3$) across all
    tool-calling accuracy metrics and outcome accuracy. Error bars and
    shaded regions indicate $\pm$1 standard deviation. All $\pm$std
    values are annotated above each bar. WS scores are scaled by
    $100$ for uniform axis range.}
  \label{fig:deepseek_errorbar}
\end{figure}
\subsection{DeepSeek-V3.2 Results with Variance}
Fig.~\ref{fig:deepseek_errorbar} reports DeepSeek-V3.2 performance
($n{=}3$ runs) for all five metrics across the three compared
configurations. Error bars and shaded regions denote $\pm$1 standard
deviation; annotated values above each bar show the exact $\pm$std for
every method--metric pair. WS scores are multiplied by 100 for a
uniform axis scale.

\subsection{Extended Evaluation on ThinkGeo with DeepSeek}
\label{sec:appendix_deepseek}

To provide a broader performance context, we extended our evaluation to the ThinkGeo dataset using the DeepSeek-V3.2 model. For this benchmark, we randomly sampled 100 questions and reformulated them into a multiple-choice format. To ensure a rigorous evaluation and eliminate heuristic shortcuts, we implemented a hint-free distractor strategy:

\begin{itemize}
    \item \textbf{Numerical:} Distractors are generated by applying a random scaling factor ($0.35\times$ to $2.2\times$) to the ground truth, ensuring they remain within a plausible magnitude.
    \item \textbf{Directional:} Substitutions are restricted to intra-category cardinal directions (e.g., East $\rightarrow$ South/North/West) to maintain natural phrasing.
    \item \textbf{Counting:} Distractors use neighboring integers (GT $\pm 1 \sim 3$) to force precise quantification.
    \item \textbf{Descriptive:} Key adjectives are replaced with direct antonyms (e.g., \textit{closer} $\leftrightarrow$ \textit{farther}).
    \item \textbf{Binary:} All four options provide substantive, content-driven answers rather than suggestive fillers like ``unknown.''
\end{itemize}

The comparative results are summarized in Table~\ref{tab:thinkgeo_acc}. The data indicates that our proposed RSMeM model achieves a significant accuracy improvement of 17.0\% over the Baseline, reaching an overall accuracy of 49.0\%.

\begin{table}[ht]
\centering
\small
\begin{tabular}{lc}
\toprule
\textbf{Model} & \textbf{Acc. (\%)} \\
\midrule
Baseline (DeepSeek-V3.2) & 32.0 \\
\textbf{RSMeM} (Ours)    & \textbf{49.0} \\
\midrule
\textbf{Net Gain}        & \textbf{+17.0} \\
\bottomrule
\end{tabular}
\caption{Accuracy comparison on the ThinkGeo subset ($n=100$) using DeepSeek-V3.2 as the backbone.}
\label{tab:thinkgeo_acc}
\end{table}

\subsection{Comparison with Retrieval Baselines}
\label{sec:appendix_retrieval_baselines}
To empirically isolate the contribution of RSMeM's structured grounding from generic retrieval augmentation, we evaluate two additional baselines on 50 EarthBench tasks using DeepSeek-V3.2 (Tab.~\ref{tab:retrieval_baselines}). \textbf{Simple RAG} retrieves the top-$k$ passages via embedding similarity from a flat corpus, while \textbf{Web Retrieval} queries Bing for task-relevant context at inference time.

\begin{table}[ht]
\centering
\footnotesize
\begin{tabular}{lccc}
\toprule
\textbf{Method} & \textbf{EM} & \textbf{WS} & \textbf{Acc. (\%)} \\
\midrule
ReAct + Simple RAG & 0.0494 & 0.2603 & 36.73 \\
ReAct + Web Retrieval & 0.0437 & 0.2603 & 48.98 \\
\textbf{RSMeM (HKG + FAR)} & \textbf{0.3584} & \textbf{0.6126} & \textbf{53.06} \\
\bottomrule
\end{tabular}
\caption{Comparison with retrieval baselines on 50 EarthBench tasks (DeepSeek-V3.2).}
\label{tab:retrieval_baselines}
\end{table}

RSMeM outperforms Simple RAG by +16.33 pp in accuracy and +0.3092 in EM, confirming that taxonomy-aware structured retrieval provides fundamentally different value than flat semantic matching. Compared to Web Retrieval, RSMeM still achieves +4.08 pp in accuracy despite Bing's access to broader information, demonstrating that curated domain knowledge with failure-aware refinement is more effective than unconstrained web context for specialized geoscientific tasks.

\subsection{Per-Domain Performance Breakdown}
\label{sec:appendix_domain_breakdown}
To understand where RSMeM provides the most benefit, we report per-domain accuracy changes using DeepSeek-V3.2 on EarthBench (Tab.~\ref{tab:domain_breakdown}).

\begin{table}[ht]
\centering
\footnotesize
\setlength{\tabcolsep}{3pt}
\begin{tabular}{lccc}
\toprule
\textbf{Domain} & \textbf{Baseline} & \textbf{RSMeM} & \textbf{$\Delta$ (pp)} \\
\midrule
Disaster Management & 48.72\% & 58.97\% & \textbf{+10.25} \\
Environmental Monitoring & 51.77\% & 58.45\% & +6.68 \\
Urban Planning \& Dev. & 53.73\% & 55.22\% & +1.49 \\
\bottomrule
\end{tabular}
\caption{Per-domain accuracy breakdown (DeepSeek-V3.2). RSMeM provides the largest gains in domains with standardized multi-step workflows.}
\label{tab:domain_breakdown}
\end{table}

RSMeM benefits most in Disaster Management (+10.25 pp), where tasks involve the most standardized multi-step workflows; HKG constraints reduce step omissions and parameter violations, while FAR guardrails curb repeated failures. Environmental Monitoring gains are moderate (+6.68 pp) due to longer pipelines with cross-modal dependencies. Urban Planning shows the smallest gain (+1.49 pp), as its shorter pipelines yield a stronger baseline and residual errors are more often dominated by upstream perception limits.
\end{document}